\documentclass[letterpaper]{article} 
\usepackage{aaai23}  
\usepackage{times}  
\usepackage{helvet}  
\usepackage{courier}  
\usepackage[hyphens]{url}  
\usepackage{graphicx} 
\usepackage{natbib}  
\usepackage{caption} 
\usepackage{algorithm}
\usepackage{algorithmic}
\usepackage{amsmath}
\usepackage{amsfonts}

\usepackage{booktabs,multirow} 

\usepackage{graphicx} 

\usepackage{newfloat}
\usepackage{listings}
\DeclareCaptionStyle{ruled}{labelfont=normalfont,labelsep=colon,strut=off} 
\floatstyle{ruled}
\newfloat{listing}{tb}{lst}{}
\floatname{listing}{Listing}
\title{Evaluating Digital Agriculture Recommendations with Causal Inference}
\author{
    Ilias Tsoumas\equalcontrib\textsuperscript{\rm 1,2},
    Georgios Giannarakis\equalcontrib\textsuperscript{\rm 1},
    Vasileios Sitokonstantinou\textsuperscript{\rm 1},
    Alkiviadis Koukos\textsuperscript{\rm 1},\\
    Dimitra Loka \textsuperscript{\rm 3},
    Nikolaos Bartsotas\textsuperscript{\rm 1},
    Charalampos Kontoes\textsuperscript{\rm 1},
    Ioannis Athanasiadis\textsuperscript{\rm 2}
}
\affiliations{
    \textsuperscript{\rm 1}BEYOND Centre, IAASARS, National Observatory of Athens\\
    \textsuperscript{\rm 2}Wageningen University and Research\\
    \textsuperscript{\rm 3}Hellenic Agricultural Organization ELGO DIMITRA\\
    \{i.tsoumas, giannarakis, vsito, akoukos, nbartsotas, kontoes\}@noa.gr\\
    \{ilias.tsoumas, ioannis.athanasiadis\}@wur.nl\\
    \{dimitra.loka\}@elgo.gr
}

\begin{document}

\maketitle

\begin{abstract}
In contrast to the rapid digitalization of several industries, agriculture suffers from low adoption of smart farming tools. Even though recent advancements in AI-driven digital agriculture can offer high-performing predictive functionalities, they lack tangible quantitative evidence on their benefits to the farmers. Field experiments can derive such evidence, but are often costly, time consuming and hence limited in scope and scale of application.
To this end, we propose an observational causal inference framework
for the empirical evaluation of the impact of digital tools on target farm performance indicators (e.g., yield in this case). This way, we can increase farmers' trust via enhancing the transparency of the digital agriculture market, and in turn accelerate the adoption of technologies that aim to secure farmer income resilience and global agricultural sustainability against a changing climate. 
As a case study, we designed and implemented a recommendation system for the optimal sowing time of cotton based on numerical weather predictions, which was used by a farmers' cooperative during the growing season of 2021. We then leverage agricultural knowledge, collected yield data, and environmental information to develop a causal graph of the farm system. Using the back-door criterion, we identify the impact of sowing recommendations on the yield and subsequently estimate it using linear regression, matching, inverse propensity score weighting and meta-learners.
The results revealed that a field sown according to our recommendations exhibited a statistically significant yield increase that ranged from $12\%$ to $17\%$, depending on the method. The effect estimates were robust, as indicated by the agreement among the estimation methods and four successful refutation tests. We argue that this approach can be implemented for decision support systems of other fields, extending their evaluation beyond a performance assessment of internal functionalities.
\end{abstract}

\section{Introduction}



The increasing global population and the changing climate are putting pressure on the agricultural sector, demanding the sustainable production of adequate quantities of nutritious food, feed and fiber. Nowadays, many industries enjoy automated and effective decision-making via harnessing the data that digitalization generates. However, the agricultural sector experiences limited adoption of precision agriculture and smart farming technologies \cite{gabriel2022adoption}. This might seem odd at first sight, given the surge of sophisticated digital tools that utilize Artificial Intelligence (AI) techniques and combine remote sensing data with data from Internet of Things (IoT) sensors to offer agricultural information of great detail \cite{sharma2020machine, nanushi2022pest, choumos2022towards}.
Yet farmers are skeptical about the effectiveness and actual contribution of these tools to their revenues and daily work \cite{lowenberg2019setting, lioutas2021digitalization}. 

Traditionally, quantifying the impact of a service would require the design and execution of a randomized experiment \cite{boruch1997randomized}. Nevertheless, field experiments for the evaluation of digital agriculture tools are seldom done since they are costly and time-consuming, requiring specialized designs and follow-up experiments for any changes in the product \cite{vaessen2010challenges, diggleexperimental}. In addition, field experiments can jeopardize the crops and hence the farmers' livelihood; and if potential damages are not covered, no prudent farmer would want to participate. As a result, the providers of digital agriculture tools often resort to unproven promises that unavoidably create customer mistrust. An observational causal inference framework \cite{pearl2009causality} can fill this gap by emulating the experiment we would have liked to run \cite{hernan2016using}.

\section{Related Work}

Causal inference with observational data has been the subject of recent work across diverse disciplines, including 
ecology \cite{arif2022utilizing}, public policy \cite{fougere2019causal}, and Earth sciences
\cite{massmann2021causal, runge2019inferring}. In agriculture, it has been used to identify and estimate the effect of agricultural practices on various agro-environmental metrics \cite{qian2016applying, deines2019satellites, giannarakis2022personalizing, giannarakis2022towards}.
Using causal inference to test digital agriculture can provide reliable insights of superior socioeconomic impact than of those inferred by naive descriptive studies, including transparent benefits for the farmers, increased reliability, and honest pricing.

According to Adelman (1992), the comprehensive evaluation of decision support systems has three facets: i) the subjective evaluation that assesses the system from the perspective of the end-user, ii) the technical evaluation that assesses the performance of the system's internal functionalities and iii) the empirical evaluation that experimentally assesses the impact of the system \cite{adelman1992evaluating}. Subjective evaluation has been widely practiced for decision support  \cite{zhai2020decision}  and recommender systems \cite{pu2012evaluating}, retrieving user feedback (e.g., on usability, accessibility etc.) through surveys and questionnaires. From the perspective of the AI-based expert systems, the technical evaluation is based on predictive, classification and ranking accuracy metrics \cite{schroder2011setting}. Technical evaluation metrics extending beyond accuracy are used within the context of recommender systems (e.g., coverage, serendipity) \cite{ge2010beyond}. These metrics capture the recommendation quality as perceived by the user, connecting the technical and subjective evaluation concepts.

Interestingly, empirical evaluation methods, and in particular with regards to the impact assessment of digital agriculture tools, have been seldom employed. 
From the perspective of agricultural economics, tangential questions have been studied \cite{muller1974sources, roberts2009estimating, schimmelpfennig2016sequential, mcfadden2022information}, but without approaching the question from a farm system standpoint, hence not leveraging available structural knowledge and reaping its benefits \cite{cinelli_crash_2020}.
Thus, we propose a framework for the empirical evaluation of digital agriculture recommendations with causal inference.
In this context, we designed and implemented a recommendation system for the optimal sowing of cotton. The system was tested in a real-world case study by providing it to a local agricultural cooperative and monitoring the results.

For several arable crops (e.g., cotton, maize, chickpea) sowing time is of great importance. Mistimed sowing can lead to suboptimal plant emergence and adversely affect the crop yield \cite{huang2016different,nielsen2002delayed,richards2022impact}. 
The agro-climatic conditions for optimal sowing  have been extensively studied \cite{freeland2006agrometeorology,boman2005soil}. Soil temperature, soil moisture and ambient temperature, during the first days after sowing, play a crucial role in proper germination and emergence and ultimately determine the yield and its quality \cite{bradow2010germination,bauer1998planting}. The University of California \cite{california} and the Texas Tech University \cite{barbato2011temperature} have developed digital tools that provide daily recommendations for optimal cotton sowing, using air temperature forecasts from the National Oceanic and Atmospheric Administration (NOAA). The accuracy of the University of California's cotton planting tool was studied by comparing the temperature forecasts to ground truth measurements from two weather stations \cite{munier2004accuracy}. The tool forecasted the correct planting conditions \cite{kerby1989weather} 75\% of the time, showing that weather forecast models can provide solid sowing recommendations.

To the best of our knowledge, there are no works that evaluate the effectiveness of any type of decision support or recommendation system in the agricultural sector through causal reasoning and 
beyond their predictive accuracy \cite{luma2020causal, pasquel2022review}. The contributions of this work are summarized as follows: i) the design and implementation of the first empirical evaluation framework for digital agriculture based on causal inference; ii) the development of a high-resolution, knowledge-based recommendation system for the optimal sowing of cotton using weather predictions, which was operationally used in a real-world case study;
iii) the identification of the causal effect of sowing recommendations on yield, its subsequent estimation with different methods (linear regression, matching, inverse propensity score weighting and meta-learners), and the use of refutation tests to evaluate the robustness of estimates. Due to its accurate weather forecasts and customized rules, we find that the system offered effective sowing recommendations. It increased the cotton yield of the farmers that
followed the recommendations by a factor that ranged from 12\% to 17\%, depending on the estimation method used.

\section{Agricultural Recommendation System}

In this work, we design, implement and evaluate a knowledge based recommendation system \cite{aggarwal2016knowledge} for optimal cotton sowing. The recommendations are based on satisfying specific environmental conditions, as retrieved from the related literature, which would ensure
successful cotton planting. The system is operationally deployed using high resolution weather forecasts. Sec. 1 of the Appendix contains an algorithmic presentation of the system.

According to literature, the minimum daily-mean soil temperature for cotton germination is $16^\circ$C \cite{bradow2010germination}. Soil or ambient temperatures lower than $10^\circ$C result in less vigorous and malformed seedlings \cite{boman2005soil}. As a general rule for cotton, agronomists recommend daily-mean soil temperatures higher than $18^\circ$C for at least 10 days after sowing and daily-maximum ambient temperatures higher than $26^\circ$C for at least 5 days after sowing. We summarize the conditions for optimal cotton sowing in Table~\ref{fig:ag_rules} \cite{freeland2006agrometeorology, boman2005soil}. Using these conditions and Numerical Weather Predictions (NWP) we implement a recommendation system that advises on whether any given day is a good day to sow or not.

\begin{table}[!ht]
\centering
\begin{tabular}{@{}lllll@{}}
\toprule
\textbf{\begin{tabular}[l]{@{}l@{}}Type of \\ Temperature\end{tabular}} &
  \textbf{Statistic} &
  \textbf{Condition} &
  \textbf{\begin{tabular}[c]{@{}l@{}}Condition\\ Priority\end{tabular}} \\ \midrule
soil (0-10 cm)  & mean & \textgreater{}$18^\circ$C    &  optimum   \\
ambient (2 m) & max  & \textgreater{}$26^\circ$C & optimum   \\
soil (0-10 cm)   & mean & \textgreater{}$15.56^\circ$C & mandatory \\
soil (0-10 cm)   & min  & \textgreater{}$10^\circ$C   & mandatory \\
ambient (2 m) & min  & \textgreater{}$10^\circ$C   & mandatory \\ \bottomrule
\end{tabular}%
\caption{Optimal conditions for sowing cotton. All conditions refer to the period from sowing day to 5 days after, except the first soil condition that refers to 10 days after. \label{fig:ag_rules}}
\end{table}

Open-access high-resolution NWP forecasts are rarely available. For this reason, we implement the WRF-ARW model \cite{skamarock2019description} with a grid resolution of 2 km. This enables us to reach a high spatio-temporal resolution for parameters that are crucial during the cotton seeding period; soil and ambient temperature are retrieved at an hourly rate for the forthcoming 2.5 days.
Ideally, 10-day predictions at a 2 km spatial resolution should be available every morning, as it is required by the conditions in Table \ref{fig:ag_rules}. However, this would demand an enormous amount of computational power. To simulate the desired data, we combine the 2.5-day high resolution forecasts with the GFS \cite{cisl_rda_ds084.1} 15-day forecasts that are given on a 0.25 degrees (roughly 25 km) spatial resolution.

\begin{equation} \label{eq:a_gfs}
a_{i}= \frac{GFS_{day=i}}{GFS_{day=1}}, i\in \{3, ..., 10\}
\end{equation}

\begin{equation} \label{eq:art}
ART_{j}=\left\{
\begin{array}{ll}
WRF_{day=j} &, j\in \left \{ 1,2 \right \} \\ 
WRF_{day=1}\cdot  a_{j} &, j\in \{3, ..., 10\}
\end{array}
\right.
\end{equation}

Eq. (\ref{eq:a_gfs}) shows how we extract the 10-day weather trend factor using GFS forecasts. We calculate the percentage change between each forecast (for $day=3$ to $day=10$) and the corresponding next day ($day=1$) forecast.
Eq. (\ref{eq:art}) shows how we produce the artificial (ART) 10-day forecasts at 2 km spatial resolution. We keep the original WRF forecasts for the next two days and for the rest we apply the respective 10-day trend factor to the next day WRF forecast.

\begin{figure}[!ht]
  \centering
  \includegraphics[scale=0.17]{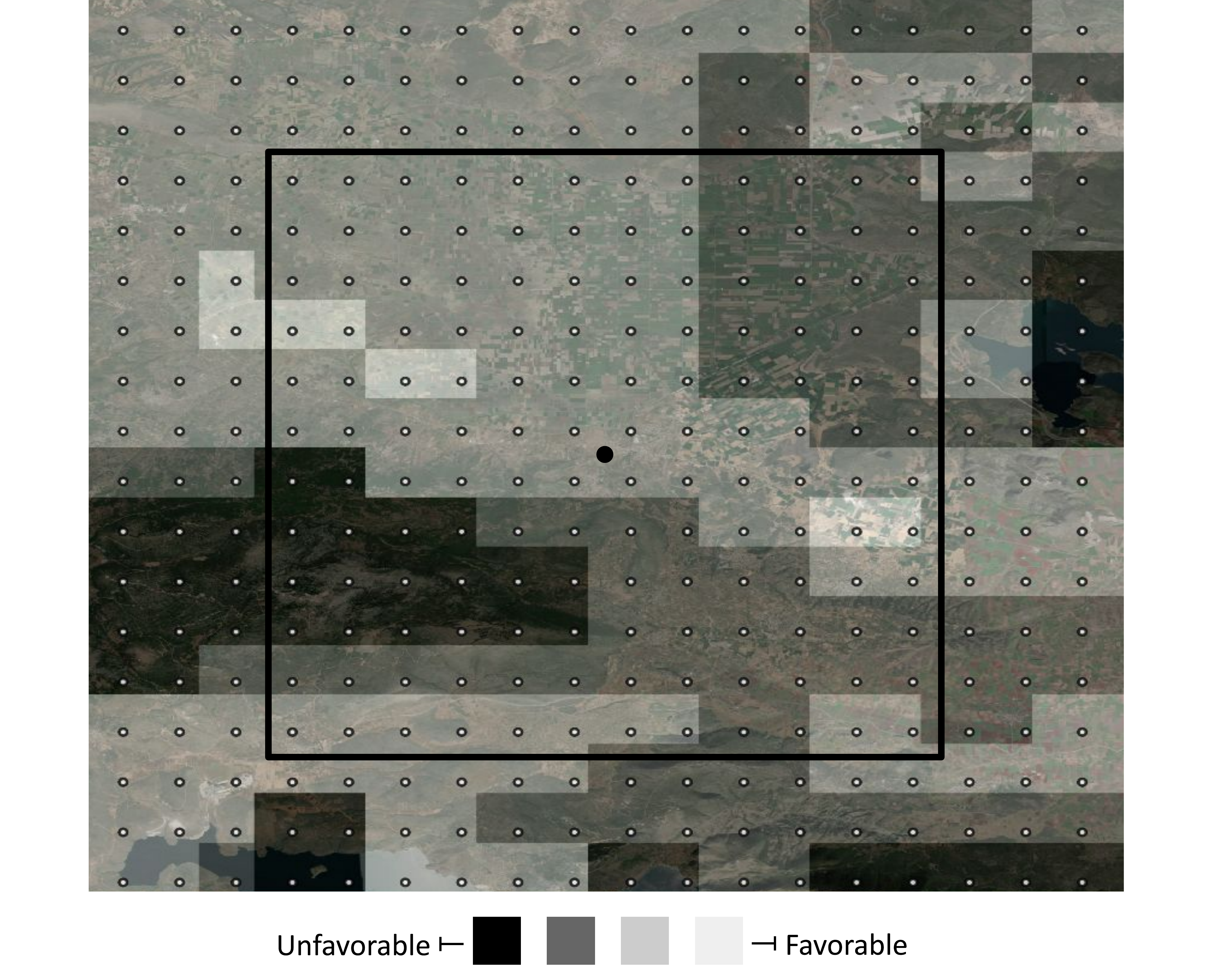}
  \caption{Optimal sowing map for a given day. The black circle at the center depicts the GFS grid point that represents the entire black-lined box. The white circles depict the 144 ART grid points for the same area.  \label{fig:spatial_gfsvsart}}
\end{figure}

We generate ART forecasts in order to provide recommendations that can vary up to the field-level, which would have been impossible with GFS forecasts alone. This is depicted in Figure \ref{fig:spatial_gfsvsart}. In order to evaluate the quality of our ART forecasts, we compared them with measurements from the nearest operational weather station in the area of interest for the critical sowing period, from 15/4/2021 to 15/5/2021. We have limited our comparison to the maximum and minimum ambient temperatures, as there were no soil temperature measurements available. It is worth noting that the nearest grid point of GFS to the station is only $0.87$ km away, however the maximum distance can be up to 12 km away. On the other hand, the equivalent grid point of ART is $1.41$ km away, which incidentally is the maximum possible distance between any location and the nearest ART point.

Initially, we compared the next day forecasts of GFS against their ART (or WRF) equivalent. The comparison analysis revealed a Mean Absolute Error (MAE), between the two forecasts and the station for maximum ambient temperature, equal to $2.39^\circ$C (GFS) versus $1.48^\circ$C (ART), and for minimum ambient temperature $1.52^\circ$C (GFS) versus $1.74^\circ$C (ART). Overall, WRF appears to behave well and slightly better than GFS. This difference is expected to be greater for other locations in the grid, as for this particular case the station happened to be very close to the GFS grid point. 
Furthermore, we calculated the MAE and Root Mean Squared Error (RMSE) of all daily 5-day forecasts of ART against the ground station for a period of interest (for graphical comparisons, see Appendix Figure $1$). For the maximum temperature we found $MAE=2.41$, $RMSE=3.11$, whereas for the minimum temperature we found $MAE=2.75$, $RMSE=3.70$.

\subsubsection{Real-World Case Study.}

We combined the ART weather forecasts and the conditions listed in Table \ref{fig:ag_rules} to produce a recommendation system in the form of daily maps over the fields of the farmers of the cooperative (Figure \ref{fig:spatial_gfsvsart}). The sowing recommendation maps were served through the website of the cooperative that farmers visited on a daily basis.
The cooperative collected and provided for each field: its geo-referenced boundaries, the sowing date, the seed variety, the harvest date, the precise final yield, and for a subset of the fields the yield of the previous year. We then combined this data with publicly available observations from heterogeneous sources, such as satellites (Sentinel-2), weather stations and GIS maps, to engineer an observational dataset that enables a causal analysis for studying the impact of the recommendation system on the yield.

\section{Causal Evaluation Framework}

\subsubsection{Notation and Terminology.}

We encode the farm system in the form of a Directed Acyclic Graph (DAG) $G \equiv (V,E)$ where $V$ is a set of vertices consisting of all relevant variables, and $E$ is a set of directed edges connecting them \cite{pearl2009causality}. The directed edge $A \rightarrow B$ indicates causation from $A$ to $B$, in the sense that changing the value of $A$ and holding everything else constant will change the value of $B$. We are using Pearl's $do$-operator to describe interventions, with $\mathbb{P}(Y=y|do(T=t))$ denoting the probability that $Y = y$ given that we intervene on the system by setting the value of $T$ to $t$. Following popular terminology, we name the variable $T$, of which we aim to estimate the effect, as \textit{treatment} and the variable $Y$, which we want to quantify the impact of $T$ on, as \textit{outcome}. The parents of a node are its \textit{direct causes}, while a parent of both the treatment and outcome is referred to as a \textit{common cause} or \textit{confounder}. Our end goal is to account for exactly the variables $Z \subseteq V$ that will allow us to estimate the Average Treatment Effect (ATE) of the treatment on outcome, as shown in Eq. (\ref{eq:ate}).
\begin{equation}\label{eq:ate}
    \text{ATE} = \mathbb{E}\big[Y|do(T=1)\big] - \mathbb{E}\big[Y|do(T=0)\big]
\end{equation}

\subsubsection{Problem Formulation.}

We thus aim to develop a causal graph $G$ whose vertices $V$ capture the relevant actors of the system we study, and edges $E$ indicate their relationships. 
The system recommendations should be part of the graph, along with cotton yield and the agro-environmental conditions that interfere in this physical process.

Because the end goal is the evaluation of the recommendation system and its actual impact on yield, we designate as \textit{treated} the fields that farmers sowed on a day that was seen as favorable by the system, i.e., the corresponding 10-day ART forecasts satisfied the appropriate conditions, and as \textit{control} the fields that were sown on a non-favorable day. Because the system outputs $4$ levels of recommendation ranging from $0$ (bad) to $3$ (good), we define a day as favorable when all conditions are satisfied, i.e., when the system outputs the highest recommendation value that is $3$. A day is then defined as non-favorable when the system outputs any other recommendation value. Binarizing the treatment in that way allows for greater flexibility in estimator selection and easier interpretation. Formally,

\[   
T = 
     \begin{cases}
       1 &\quad\text{, if recommendation of sowing date} \in $ \{3\}$\\
       0 &\quad\text{, if recommendation of sowing date} \in $ \{0,1,2\}$  
     \end{cases}
\]

Beyond the recommendation system, multiple factors influence the decision to sow or not. This is precisely the challenge we aim to address by employing a graphical analysis and explicitly modeling the farm system structure. The ATE we aim to estimate captures the difference between what the average yield would have been if we intervened and forced farmers to follow the recommendation by sowing on a favorable day, and the average yield if we forced them to defy the recommendation by sowing on an unfavorable day. Such an estimand is of primary significance for the farmers, but also for proving the reliability and therefore accelerating the adoption of smart farming tools. 
Given that 
confounding factors are controlled for, we henceforth refer to the ATE as the \textit{(average) causal effect of following the recommendation} in the sense described above.

\subsubsection{Cotton Domain Knowledge and Graph Building.}

Cotton yield and quality are ultimately determined by the interaction between the genotype, environmental conditions and management practices throughout the growing season. Nevertheless, the first pivotal steps for a profitable yield are a successful seed germination and emergence which are greatly dependent on timely sowing \cite{wanjura1969emergence, bauer1998planting,  bradow2010germination}.

Emergence and germination mediate the effect of $T$ on $Y$; however, Crop Growth ($CG$) was not observed. We thus turned to the popular Normalized Difference Vegetation Index (NDVI) in order to obtain a reliable proxy of $CG$, and specifically used the trapezoidal rule across NDVI values from sowing to harvest 
\cite{eklundh2015timesat}. Even though in the case of cotton, trapezoidal NDVI is not linearly correlated with yield \cite{dalezios2001cotton, zhao2007canopy}, it is correlated with early season Leaf Area Index (LAI) \cite{zhao2007canopy}, which in turn is a good indicator of early season crop growth rate \cite{virk2019physiological}.
Furthermore, seed germination and seedling emergence are greatly dependent on soil moisture. Hence, soil moisture $SM$ is a confounder for the relation $T \rightarrow Y$ that we study. As a $SM$ proxy, we used the Normalized Difference Water Index (NDWI) at sowing day which is highly correlated with soil moisture in bare soil \cite{casamitjana2020soil}. 

Agricultural management practices before sowing ($AbS$) comprise tilling operations for preparing a good seedbed. Practices during sowing ($AdS$) include a sowing depth of $4-5$ cm and an average distance of $0.91$ m between rows and $7.62$ cm between seeds. After sowing practices ($AaS$) comprise basic fertilization, irrigation and pest management. It is reasonable to think that all aforementioned practices are a result of a common cause that we can define as Agricultural Knowledge ($AK$), capturing the skills and experience of a farmer. We possess no quantitative information on the agricultural knowledge or the practices followed by each farmer. However, the farmer's cooperative is not large, and aims for consistent, high-quality produce. As a result, they have developed highly consolidated routines for interacting with their crops: this includes common practices, homogeneous fertilizer application, and jointly owned machinery. We thus note that even if we do not have numerical data on $AbS, AdS, AaS$, the cooperative directors do not observe significant differences across fields and for the purposes of our study these variables are considered to be constant. 

At the same time, it is rational to assume that the agricultural knowledge ($AK$) of any farmer interacts with crops exclusively through management practices. Because of the aforementioned condition, the influence of $AK$ on the system is nullified and we hence omit it from the graph. While we note that the above limit the external validity of our results \cite{calder1982concept}, by assuming that agricultural practices are constant for all farmers and that $AK$ only interacts with the system through them, we implicitly control for all of them \cite{huntington2021effect}.

Apart from soil moisture, soil and ambient temperatures at the time of sowing and for 5-10 days after, affect seed germination, seedling development and final yield \cite{virk2019physiological,boman2005soil,varcosoil}.
Low temperatures result in reduced germination, slow growth and less vigorous 
seedlings that are more prone to diseases and sensitive to weed competition \cite{wanjura1969emergence,bradow2010germination}. This knowledge is incorporated in the sowing recommendations, in the form of numerical rules, and consequently in the treatment $T$. We thus added in the graph the weather forecast $WF$ (variables listed in Table \ref{fig:ag_rules}) as a parent node of $T$. We also had access to the weather on the day of sowing $WS$ (min \& max ambient temperature in $^\circ$C) from a nearby weather station, influencing $WF$, $T$, and $CG$.

Topsoil (0-20 cm) properties $SP$ (\% content of clay, silt and sand) and organic carbon content  $SoC$ (g C kg\textsuperscript{-1}) also affect cotton seed germination and seedling emergence due to differences in water holding capacity and consequently in soil temperature and aeration, drainage and seed-to-soil contact \cite{varcosoil}. Data on $SP$ and $SoC$ were retrieved from the European Soil Data Centre (ESDAC) \cite{ballabio2016mapping, de2015map}. Both variables were included in the graph as confounders of $T$ and $CG$.
Seed variety also determines seed germination, emergence and final yield \cite{sniderseed}.
Seed mass and vigor \cite{liu2015early,sniderseed} are related to the seed variety ($SV$); we hence added the latter as a confounder for $T$ and $Y$. In this case, we had 13 different cotton SVs.

\begin{table}[!ht]
\small
\centering
\begin{tabular}{lll}
\toprule
\textbf{Id} & \textbf{Variable Description}           & \textbf{Source}         \\ \midrule
T   & Treatment & Farmers' Cooperative, RS    
\\
WF  & Weather forecast             & GFS, WRF                \\
WS  & Weather on sowing day        & Nearest weather station \\
WaS & Weather after sowing         & Nearest weather station                  \\
CG  & Crop Growth                  & NDVI via Sentinel-2                      \\
SM  & Soil Moisture on sowing      & NDWI via Sentinel-2     \\
SP  & Topsoil properties           & Map by ESDAC \\
SoC & Topsoil organic carbon       & Map by ESDAC \\
SV  & Seed Variety                 & Farmers' Cooperative \\
G   & Geometry of field           & Farmers' Cooperative                  \\
AdS & Practices during sowing      & Farmers' Cooperative \\
AbS & Practices before sowing      & Farmers' Cooperative \\
AaS & Practices after sowing       & Farmers' Cooperative \\
HD  & Harvest Date                 & Farmers' Cooperative \\
Y   & Outcome (Yield)              & Farmers' Cooperative \\ \bottomrule
\end{tabular}%
\caption{Farm system variable identifier, description and source. RS = Recommendation System.}
\label{tab:variables}
\end{table}

The geometrical properties of the field (perimeter to area ratio, $G$) were also considered, as border effects can play a minor role on crop growth, confounding the effect of $T$ on $Y$
\cite{green1956border}. Since temperature is the primary environmental factor controlling plant growth \cite{bange2004impact,hatfield2015temperature}, temperature fluctuations were observed throughout the growing season from the nearest weather station, constituting a parent variable $WaS$ (min \& max ambient temperature in $^\circ$C) of crop growth $CG$. 
Lastly, the Harvest Date ($HD$) mediates the effect of $CG$ on $Y$, influencing both yield potential and quality \cite{dong2006yield,bange2008managing}. Table \ref{tab:variables} summarizes the variables' description, abbreviation and source.  

\subsubsection{Causal Graph.}

\begin{figure}[!ht]
  \includegraphics[scale=0.166]{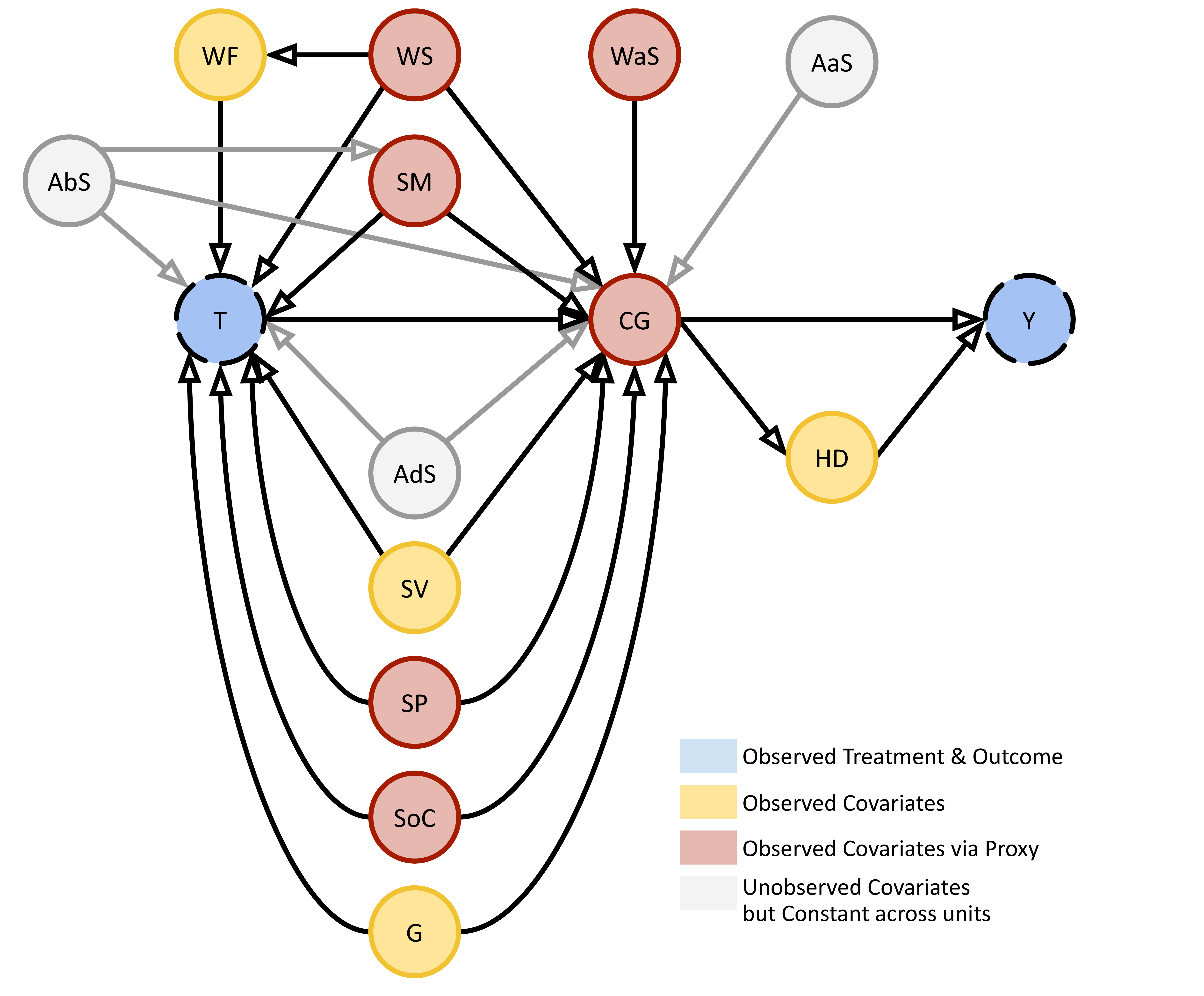}
  \caption{Graph of the farm system, encoding the causal relations between the relevant agro-environmental actors. \label{fig:causal-graph}}
\end{figure}

Figure \ref{fig:causal-graph} displays the final causal graph $G$. We note that, in reality, it is impossible to account for all factors interacting in the system in order to claim that the estimated effect will not contain any bias. However, because the selection of variables is deeply rooted on well-understood agro-environmental interactions, bias is expected to be minimized, in the sense that no important interactions are left unaccounted for. Furthermore, we extensively test the reliability of effect estimates through multiple refutation checks.

\subsubsection{Effect Identification and Estimation.}

Because the calculation of causal effects requires access to counterfactual values that are by definition not observed \cite{holland1986statistics}, observational methods rely on identification techniques and assumptions that aim at reducing causal estimands such as $\mathbb{P}(Y=y|do(T=t))$ to statistical ones, such as $\mathbb{P}(Y=y|T=t)$. The back-door criterion is a popular identification method that solely relies on a graphical test to infer whether adjusting for a set of graph nodes $Z\subseteq V$ is sufficient for identifying $\mathbb{P}(Y=y|do(T=t))$ from observational data. Formally, a set of variables $Z$ satisfies the back-door criterion relative to an ordered pair of variables $(T, Y)$ in a DAG $G$, if no node in $Z$ is a descendant of $T$ and $Z$ blocks every path between $T$ and $Y$ that contains an arrow into $T$.
After (if) we have obtained a back-door adjustment set to condition on, we can proceed with estimating the ATE of interest. The back-door criterion already provides a formula for the interventional distribution. Given a set of variables $Z$ satisfying the back-door criterion we can identify the causal effect of $T$ on $Y$ as $\mathbb{P}(y|do(t)) = \sum_z \mathbb{P}(y|t, z)\mathbb{P}(z)$. 

In our study, ATE estimation is done with several methods of varying complexity. To check covariate balance and as a method prerequisite, we model the propensity scores $\mathbb{P}(T=1|Z=z)$, i.e., the probability of receiving treatment given features \cite{rosenbaum1983central}. Linear regression and distance matching are selected as baseline estimation methods. The popular Inverse Propensity Score (IPS) weigthing is also used \cite{stuart2010matching}. We finally apply modern machine learning methods, i.e., the baseline T-learner and the state-of-the-art X-learner \cite{kunzel_metalearners_2019}.

\subsubsection{Refutation Methods.}

One of the biggest challenges in causal inference pertains to model evaluation. Given the fact that ground truth estimates are not observed, we resort to performing robustness checks and sensitivity analyses of estimates, in line with recent research \cite{sharma2020dowhy, cinelli2020making}. We perform the following tests: i) Placebo treatment, where the treatment is randomly permuted and the estimated effect is expected to drop to $0$; ii) Random Common Cause (RCC), where a random confounder is added to the dataset and the estimate is expected to remain unchanged; iii) Random Subset Removal (RSR), where a subset of data is randomly selected and removed and the effect is expected to remain the same; iv) Unobserved Common Cause (UCC), where an unobserved confounder acts on the treatment and outcome without being added to the dataset, and the estimates should remain relatively stable. The Placebo, RCC and RSR tests are bootstrapped to generate confidence intervals and p-values \cite{diciccio1996bootstrap}. The UCC returns a heatmap of new ATE estimates depending on the strength of unobserved confounding.

\section{Experiments, Results and Discussion}

\begin{table*}[!ht]
\small
\centering
\begin{tabular}{cccccccccccc}
\toprule
\multicolumn{4}{c}{\multirow{2}{*}{\textbf{Causal Effect Estimation}}} & \multicolumn{8}{c}{\textbf{Refutations}}                                              \\ \cmidrule(l){5-12} 
\multicolumn{4}{c}{} &
  \multicolumn{2}{c}{\textbf{Placebo}} &
  \multicolumn{2}{c}{\textbf{RCC}} &
  \multicolumn{2}{c}{\textbf{UCC}} &
  \multicolumn{2}{c}{\textbf{RSR}} \\ \midrule
\textbf{Method} & \textbf{ATE} & \textbf{CI} &  \textbf{p-value} 
& \textbf{Effect*} & \textbf{p-value} & \textbf{Effect*} &
  \textbf{p-value} &   \multicolumn{2}{c}{\textbf{Effect*}} &
  \textbf{Effect*} & \textbf{p-value} \\ 
  \midrule
\textbf{Linear Regression}   & 546   & (211, 880)   & 0.0015  & -25.74 & 0.39 & 546 & 0.49  & \multicolumn{2}{c}{85} & 543 & 0.45 \\
\textbf{Matching}            & 448   & (186, 760)   & 0.0060   & 50.82 & 0.39 & 432  & 0.40 & \multicolumn{2}{c}{116} & 438 & 0.48 \\
\textbf{IPS weighting} &  471 &  (138, 816) &  0.0010 &  38.82 &  0.40 &  470 &  0.40 &  \multicolumn{2}{c}{113} &  462 &  0.45 \\
\textbf{T-Learner (RF)}      & 372   & (215, 528)   & 0.0240 & 9.26  & 0.49 & 373 & 0.46 & \multicolumn{2}{c}{-} & 353 & 0.42 \\
\textbf{X-Learner (RF)}      & 437   & (300, 574)   & 0.0050 & 5.10   & 0.50 & 430 & 0.37 & \multicolumn{2}{c}{-} & 409    & 0.36 \\ \bottomrule
\end{tabular}
\caption{Results of Average Treatment Effect estimation. Includes point estimates, $95\%$ confidence intervals, and four refutation tests. For the Placebo, RCC and RSR refutations, the new ATE estimate is reported (denoted as Effect*), alongside the respective p-value ($<0.05$ indicates a failed test). The UCC column reports the mean ATE estimate of the corresponding heatmap (for full heatmaps and details see Sec. 2 of Appendix). Numbers are in cotton kg/ha, rounded to the nearest integer.}
\label{tab:results}
\end{table*}

The sowing period lasted from early April to early May 2021, the harvest took place from mid to late September, and the yield per hectare ranged from $1,250$ to $6,960$ kg (Figure 2 of Appendix contains relevant histograms).
The dataset consists of $171$ fields ($51$ treated and $120$ control). 
Variables that registered intra-field values (NDVI, NDWI) were averaged at the field-level. For the experiments, we are using the popular doWhy \cite{sharma2020dowhy} and Causal ML \cite{chen2020causalml} Python libraries.

Applying the back-door criterion on graph $G$ (Figure \ref{fig:causal-graph}), the following adjustment set of nodes $Z$ was found to be sufficient for identifying the ATE: 

\begin{equation}
\begin{aligned}
Z = & \{\textsc{ws\textsubscript{min, max}, soc, sm, g, } \\
    & \textsc{  sp\textsubscript{silt, clay, sand}, abs, ads, sv\textsubscript{1-13}} \}
\end{aligned}
\end{equation}

\noindent All variables in $Z$ are numerical, including the one-hot encoded vectors of the categorical \textsc{SV}\textsubscript{1-13} variable of seed variety. $AbS$ and $AdS$ are constant, and thus excluded for estimation purposes. We scale the data by subtracting the mean from each variable and dividing by its standard deviation. The treatment $T$ is binarized, with $1$ indicating that a farmer sowed on a favorable day, and $0$ indicating the opposite.

\begin{figure}[!ht]
  \centering
  \includegraphics[scale=0.225]{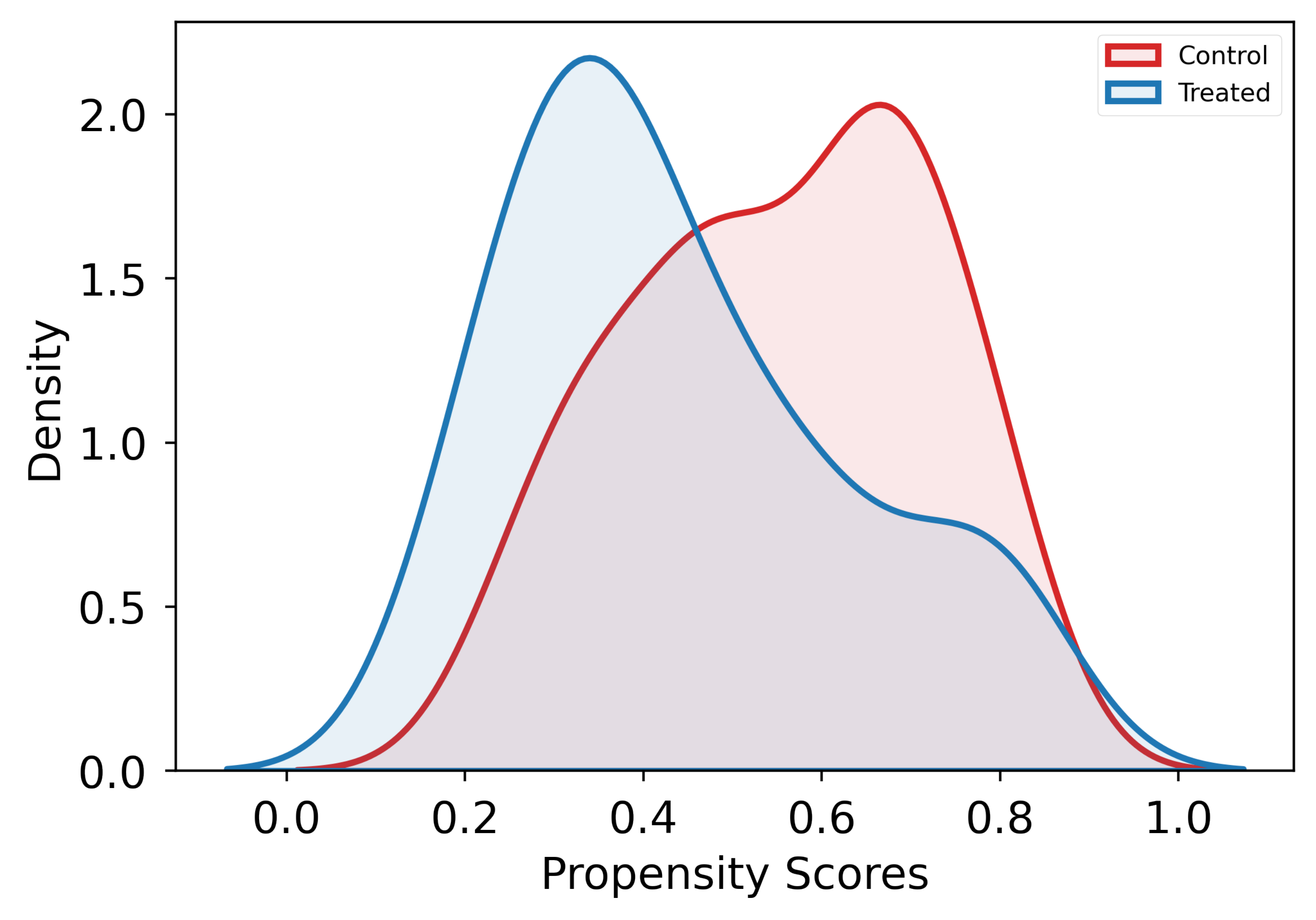}
  \caption{Distribution of propensity scores for the control and treatment group after trimming extreme scores.} \label{fig:prop_distr_trimmed}
\end{figure}

Propensity modeling is a prerequisite of IPS weighting. We thus begin by discussing the propensity model that is fit. Given the relatively small dataset size, 
logistic regression is used on the scaled back-door adjustment set $Z$ for classifying each field into the treatment/control group. We subsequently trim the dataset by removing all rows with extreme propensity scores ($<0.2$ or $>0.8$) to aid the overlap assumption 
\cite{imbens2015causal}. The resulting distribution of propensity scores can be seen at Figure \ref{fig:prop_distr_trimmed}. The model scores $0.81$ in accuracy, $0.64$ in F1-score, and $0.88$ in ROC-AUC.
After trimming extreme propensity scores, a subset of $48$ treated and $37$ control units remains. There is decent overlap between the propensity score distributions of the treatment and control group, indicating that they are comparable and enabling reliable propensity-based ATE estimation.

\begin{figure}[!ht]
  \includegraphics[scale=0.30]{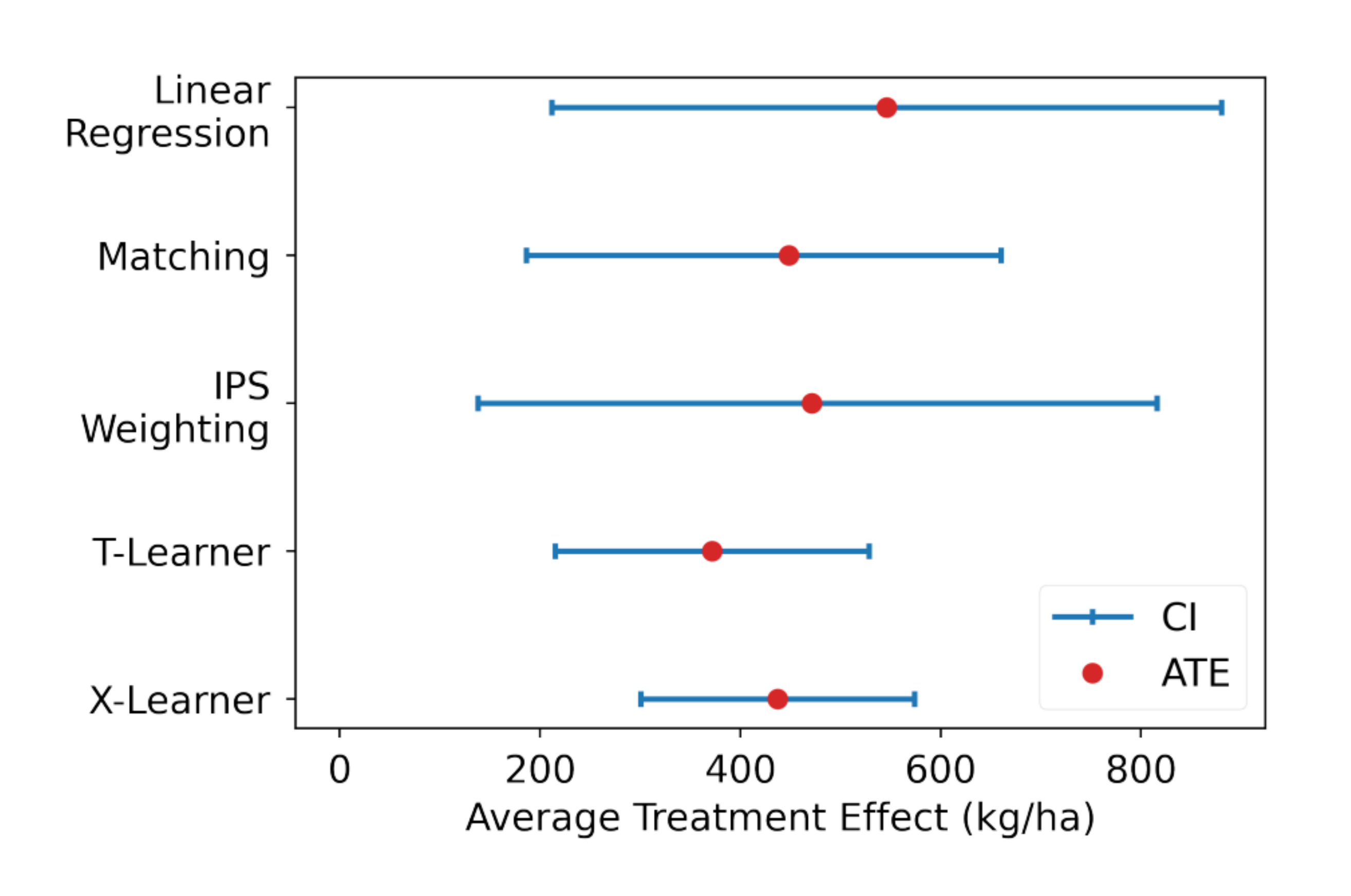}
  \caption{ATE point estimates and 95\% confidence intervals for all estimation methods.} \label{fig:boxes}
\end{figure}

Table \ref{tab:results} and Figure \ref{fig:boxes} show the results of the ATE estimation per method, alongside the corresponding $95\%$ confidence intervals and p-values. Besides Linear Regression, other methods do not provide confidence intervals by default. For matching, IPS, and meta-learners confidence intervals and the resulting p-values are hence bootstrapped. Both the T-learner and X-learner use a Random Forest for modeling the outcome $Y$. 

All methods detect a significant ATE at $95\%$ confidence level, with point estimates ranging from $372$ to $546$ kilograms of cotton per hectare. For context, the average observed yield is $3,145$ kg/ha. We thus infer that the causal effect of following the sowing recommendation on yield is significantly positive, driving a yield increase ranging from $12\%$ to $17\%$, depending on the estimation method used.

Of central importance are the refutation tests we run after having estimated the recommendation impact. Table \ref{tab:results} features analytic results for all method / refutation test combinations. All estimation methods are robust against performing the following data manipulations and re-estimating the ATE: randomly permuting the treatment (Placebo test), adding a confounder (RCC test), sampling a subset of data (RSR test) and creating unobserved confounding (UCC test). Specifically, Placebo ATE estimates do not differ significantly from $0$, while RCC and RSR estimates do not differ significantly from the already obtained ATE. For the UCC test, the mean ATE estimates are reduced yet remain positive, despite unobserved confounding of significant magnitude. Confidence intervals and p-values are bootstrapped ($1000$ iterations).

The results indicate that the recommendation system's advice drove a net increase in yield that was deemed both significant and robust from a statistical perspective. By utilizing the theory of graphical causal models, the analysis transparently puts forward its assumptions and explicitly incorporates domain knowledge in it. Combined with accurate and performant systems, such analyses can benefit the reliability and adoption of digital agriculture as well as farmers' trust. The provision of information on the actual impact expected from a recommendation system may also enable a cost-benefit analysis on behalf of the farmer, by simply comparing the digital tool cost to the expected yield gain.

Even though the analysis is transparent, it is as good as the causal assumptions it makes and the DAG it develops. Our graph is consistent with agri-environmental knowledge on cotton, however there is always a possibility that bias exists, either due to a missed confounder or due to a missed interaction between observed variables. The robustness checks we performed were all successful; noting that when we add strong unobserved confounding, the UCC test estimates become volatile - an expected behavior to a certain degree.

Given the homogeneous management practices among farmers in our data, we remark that external validity of estimates is low, as results cannot be expected to generalize to other farms that might follow different routines. Nevertheless, it is not uncommon for farmers to follow similar practices in other regions or even entire countries. While the transfer of effect estimates warrants caution, the same does not hold for the proposed framework itself. Given relevant data and knowledge, a graph-based empirical evaluation of an agricultural recommendation system can normally proceed. If consensus among estimation methods in terms of ATE significance is reached, the tool is deemed beneficial; otherwise, more work is required.
All in all, this system equips farmers with a provably valuable tool based on cotton knowledge and weather forecasts. It contributes to a successful growing season and lowers the likelihood of farmers resorting to expensive actions, e.g., replanting a field.

For the growing season of 2022, the recommendation system was deployed at national scale and extended to two other crops (maize, sunflower).
These new pilot applications will allow us to practically test the external validity of our results across different seasons, crops and locations. Moreover, given the developed causal graph $G$, the crop growth ($CG$) variable that is sufficiently captured through its NDVI proxy, mediates the effect of $T$ on $Y$. The front-door criterion \cite{pearl2009causality} might thus provide an alternative identification method for the ATE, and we plan on exploring it in collaboration with domain experts. Finally, the more growing seasons the recommendation system has seen, the more data are obtained. Going beyond ATE estimation by learning Conditional ATEs and using causal machine learning methods for providing personalized effect estimates is another next step. Due to the rich and well-established domain knowledge, we finally believe that the potential of causal reasoning in agriculture extends far beyond effect identification. Fitting Structural Causal Models and performing counterfactual inference can enable a greater understanding of the farm system and supercharge decision support tools.

Most generally, the essential condition that allowed us to utilize causal inference for empirically evaluating an agricultural recommendation system is the ample, long-established domain knowledge that exists. Decision support systems are being used on multiple fields \cite{marakas2003decision} such as medical decision making \cite{sutton2020overview} or forest and fire management \cite{segura2014decision, martell2015review}. The aforementioned fields possess accumulated domain knowledge on the interactions a good system exploits; the same way we possess information on environmental conditions related to cotton planting. We thus expect graphical approaches to be valuable for the empirical evaluation of decision support systems of diverse domains.

\section{Conclusion}

In this study, we design, implement, and test a digital agriculture recommendation system for the optimal sowing of cotton. Using the collected data and leveraging domain knowledge, we evaluate the impact of system recommendations on yield. To do so, we utilize and propose causal inference as an ideal tool for empirically evaluating decision support systems. 
This idea can be upscaled to other digital agriculture tools as well as to different fields with well-established domain knowledge. This paradigm is in principle different to decision support systems that frequently use black-box algorithms to predict variables of interest, but are oblivious to the evaluation of their own impact. In that sense, this work comes to the defence of the farmer, by introducing an AI framework for elaborating on the assumptions, reliability, and impact of a system before discussing service fees.

\section{Acknowledgements}

We thank the Agriculture Cooperative of Orchomenos for the collaboration and data provision, and Corteva Hellas for their support. This work was supported by the EU H2020 Research and Innovation program through the eshape project (grant agreement No. 820852). It was also supported by the MICROSERVICES project (2019-2020 BiodivERsA joint call / BiodivClim ERA-Net COFUND programme, and with GSRI, Greece - No. T12ERA5-00075).

\fontsize{9.0pt}{10.0pt} \selectfont
\bibliography{aaai23}

\end{document}